\documentclass{article}
\usepackage{fullpage}
\usepackage{graphicx}
\usepackage{booktabs}
\usepackage{amsmath}
\usepackage{amsthm}
\usepackage{parskip}
\usepackage{hyperref}
\usepackage{pdfpages}

\usepackage[british]{babel}

\usepackage{amsfonts}

\title{Can Large Language Models Create New Knowledge for Spatial Reasoning Tasks?}
\author{Thomas Greatrix, Roger Whitaker, Liam Turner, Walter Colombo}
\date{March 2024}

\begin{document}

\maketitle

\begin{abstract}
    The
    potential 
    for Large Language Models (LLMs) to generate new
    information 
    offers a potential step change for research and innovation. This is challenging to assert as it can be difficult to determine what an LLM has previously seen during training, making ``newness'' difficult to substantiate. In this paper we
    observe 
    that LLMs are able to perform sophisticated reasoning on problems with a spatial dimension, that they are unlikely to have previously directly encountered. While not perfect, this points to a significant level of understanding that state-of-the-art LLMs can now achieve, supporting the proposition that LLMs are able to yield significant emergent properties. In particular, Claude 3 is found to perform well in this regard. 
\end{abstract}


\section{Introduction}
Large Language models (LLMs) based on transformers have made a significant impact over just a few years. Particular surprise has concerned their capabilities as the models have progressively increased in scale. LLMs are now comparable with the brain sizes of small animals. While GPT-3's architecture exceeded the brain size of a rat with an estimated $10^{11}$ parameters \cite{mann2020language}, GPT-4 is an order of magnitude larger still, with over $10^{12}$ parameters. This is a small step closer to the human brain, having around $10^{14}$ parameters \cite{pakkenberg2003aging}. 
When combined with 
large-scale 
online training through the Internet and worldwide web, these models have surpassed expectations on the level of intelligence that a machine can achieve. Indeed, despite limitations, ``sparks of artificial general intelligence'' are reported across a range of domains, including areas of mathematical reasoning \cite{bubeck2023sparks}.

Across multiple domains, a main point of interest concerns so-called {\em emergent properties} \cite{wei2022emergent} that models seemingly possess, representing capabilities in which they haven't been explicitly trained. This is something that is not currently understood, and ripe for investigation. The extent and nature of general capabilities have also sought to divide the AI community \cite{michael2022nlp, mitchell2023debate, dell2023navigating}, 
particularly concerning i) the level of understanding that LLMs possess, and ii) the extent of their creative abilities beyond their auto-regressive training \cite{bender2021dangers, shanahan2024talking, GeoffreyHinton, yadlowsky2023pretraining, bubeck2023sparks, GuillaumeVerdon, BenBlaiszik}. 

Our aim in this paper is to report further insights as to where LLMs can create new knowledge beyond their training. Spatial reasoning problems are 
used as the basis for investigation here 
because these offer multidimensional challenges that rapidly become complex. 
The problems identified here are intentionally obscure, to ensure limited related literature. 
While overall performance in the tasks provided across the different LLMs is highly variable with some degree of hallucination, we find that Claude 3 
provides notable results that 
show how LLMs can directly support new knowledge creation, alongside supporting researchers in abstract ideation.

\section{Background}

The question of whether LLMs can consistently create 
new information which is 
both genuinely new and useful is a topic of ongoing debate within the AI research community. Many claim that LLMs are empty``stochastic parrots" \cite{bender2021dangers,JohnNaughton}
and are simply combining information from their training dataset in the pursuit of next word prediction \cite{shanahan2024talking}. 
This would limit LLMs to at best combinatorial creativity 
and restrict their ability to generalise.
Furthering these claims, recent work proposed that Transformer based models were unable to generalise beyond their training dataset \cite{yadlowsky2023pretraining}. This being said, others have noted caution on being dismissive of what LLMs understand and can achieve as they scale. The neural network pioneer Geoffrey Hinton is one such expert, who in his recent lecture series {\em ``Will digital intelligence replace biological intelligence?''} \cite{GeoffreyHinton}, 
proposes that the latest LLMs deeply understand by virtue of their underlying model. Hinton argues that the billions of features and their interaction constitutes a sophisticated model for understanding, such that emergent properties are evident. To exemplify the ability to reason beyond training, Hinton uses a simple puzzle that was put to GPT-4, before it could access the Internet. Specifically: 

{\em “The rooms in my house are painted blue or white or yellow. Yellow paint fades to white within a year. In two years time I want them all to be white. What should I do and why?”} 

In response, GPT-4 replied:

{\em “Assuming that blue paint does not fade to white over time and you want all rooms to be white in two years, here’s what you should do:
\begin{enumerate}
\item 	The rooms painted in white: You don’t have to do anything to these rooms since they are already white. 
\item The rooms painted in yellow: You don’t need to repaint these rooms either, because as per your information, yellow paint fades to white within a year. So in two years, they will be white. 
\item The rooms pained in blue: These rooms need to be repainted with white paint because the blue paint will not fade to white on its own.”
\end{enumerate}}

This is a scenario likely unseen in training data that GPT-4 navigated with ease. And moreover, the approach to reasoning is  elegant, with GPT-4 taking a mathematical reductionist approach that optimises for resources. This is an example of a powerful emergent property. Although somewhat trivial, it shows the possibility for an LLM to create new and useful ideas beyond its training.
Where direct connections to training data cannot be made in support of problem solving, the scale of data on which LLMs are trained means that they have likely seen similar or related issues to those on which they are being queried. This allows the LLM to draw on useful knowledge that maybe beyond the user's awareness.


Through such processes there is increasing evidence that LLMs can aid the creation of genuinely new 
information. 
Already LLMs have been used within larger systems in order to obtain new and/or useful artifacts. Notable examples of this include FunSearch \cite{romera2024mathematical}, which used LLMs in an evolutionary-style algorithm in order to lower the bound of the capset problem; the usage of LLMs as a high-level game designer \cite{anjum2024ink}; and the usage of LLMs in social science to perform thematic analysis \cite{tornberg2023use,dai2023llm,de2023can}. The impact of LLMs can also been seen over different stages in the research process \cite{ziems2024can,picard2023concept} and over a wide variety of fields \cite{oniani2024emerging,stella2023can,ziems2024can}. While LLMs in such instances are not performing academic research on their own, as LLM capabilities improve and autonomous agent systems like AutoGPT \cite{Significant_Gravitas_AutoGPT} and GPTdev \cite{qian2023communicative} improve, we might see LLMs being able to heavily support or automate some research tasks such as hypothesis generation.

Most recently, after the release of Claude 3, it was claimed that Claude 3 was able to reproduce unpublished results \cite{GuillaumeVerdon} and provide valuable insights on unpublished results \cite{BenBlaiszik}. Among the most notable critiques are claims that both may have been in the training dataset, however it is hard to know for sure as the dataset is proprietary and it is subject to the precise cutoff point which matters in both cases\footnote{This wasn't the only critique, but was perhaps the most common one.}. Even if both results were in the training datasets, the retrieval and recreation of these from the massive data on which Claude 3 was trained on is still quite impressive.

\subsection{Approach}
Based on the current early-stage progress in this new field, it remains important to further understand how LLMs can tackle problems beyond those seen from within their training data. This governs the extent that LLMs can significantly add to human knowledge and accelerate research and innovation.
Accordingly, we address two test scenarios that require a sophisticated spatial understanding in a mathematical context. These involve a decidable game (Section \ref{section:decidable_game}) and polygons (Section \ref{section:polygons}). These are presented to different LLMs, namely ChatGPT-3.5-Turbo, Claude 3, and Bing Copilot, which were at the time of writing state of the art. We consider their respective performance in terms of being able to make new assertions against particular spatial problems.


\section{Finding the winner of a decidable game} \label{section:decidable_game}

To explore whether LLMs can create new knowledge beyond their training, the first example we consider comes from combinatorial game theory, with the game originally coming from a TikTok video \cite{JustinFlom_2023}. The game is played as follows:

\begin{itemize}
    \item There are $n$ empty spaces in a row, and 2 players: $A$ and $B$. Player $A$ goes first.
    \item The players take it in turns to place a counter into one of the empty spaces on the line. They are not allowed to place a counter directly next to another counter, regardless of if it is their counter or not.
    \item The winner is the last person able to place a counter whilst respecting the above rules.
\end{itemize}

We believe there is no academic literature on this game - though it is worth noting that there are papers on similar topics. As an example, there is a paper on optimally picking a urinal such that the chances of having an adjacent urinal being occupied is minimised  \cite{kranakis2010urinal}. While this presentation of scenario is somewhat sensational, the general principles of the game can be more widely applied, such as in choosing parking spaces for example.

If both players play this game optimally, it is decidable by the Zermelo Theorem \cite{schwalbe2001zermelo} and as such an LLM should know who will win for a given number of empty spaces $n$. Below we present the message used to prompt a range of alternative LLMs wherein we ask about the case when $n = 7$.

\hrulefill

\textit{Two players, A and B, play a game. There is a line of ``n" empty spaces. The players take it in turns placing a counter into one of the empty spaces on the line. The players are not allowed to place a counter directly next to another counter, regardless of if it is their counter or not. The winner of the game is the last player to place a counter, making the objective of the game to make it so your opponent cannot place a counter. If both players play optimally, and player A goes first, who will win if the line has 7 spaces?}

\hrulefill

\subsection{LLM Responses}

Of the LLMs asked, Claude 3 did particularly well. Not only was it able to answer the question correctly, by stating that Player $A$ would win, Claude 3 was also able to come up with a new, provably dominant strategy for if there is an odd number of spaces. When Claude 3 was asked if it recognised the game, it responded saying that the game was the ancient game of Nim \cite{bouton1901nim}. Whilst the statement is incorrect, it is worth noting that we can prove the game is equivalent to Nim via the Sprage-Grundy Theorem \cite{grundy1939mathematics}.

GPT-3.5-Turbo and Bing Copilot, meanwhile, were not able to correctly answer who would win, and whilst both LLMs had a good attempted to find optimal strategies, the strategies proposed by GPT-3.5-Turbo and Bing Copilot did not work in general. Bing Copilot did, notably, also recognised the game as Nim however.

A full breakdown of the strategies, alongside conversation screenshots, is available and can be found in Appendix \ref{sec:Claude_response_game} to \ref{sec:GPT_response_game}.




\section{Polygons with special properties}
\label{section:polygons}

For our next task, we asked GPT-3.5-Turbo, GPT-4-Turbo, and Claude 3 about a family of polygons with special properties via the following prompt:

\hrulefill

\textit{Consider the family of polygons with 24 sides such that every side is the same length and every angle is either 90 or 270 degrees. What properties must the polygons in this family have?}

\hrulefill

\begin{figure}[h]
\centering
\includegraphics[scale = 0.6]{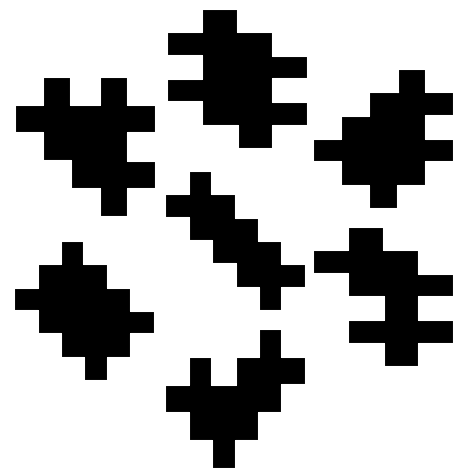}
\caption{All 7 polygons satisfying the prompt.}
\label{fig:polygon family}
\end{figure}

Whilst there is literature related to this problem, predominantly from the mathematical field of polyominos \cite{golomb1996polyominoes} (of which our family of polygons is a special case), we do not believe there to be any academic literature that directly discusses this family of polygons. In Figure \ref{fig:polygon family}, all 7 of the polygons which satisfy the prompt are exhibited.

Note that the LLMs were only provided with the prompt specified earlier, and were not provided any images of the polygons in Figure \ref{fig:polygon family}.

\subsection{Polygons with special properties results.}

Both Claude 3 and Bing Copilot proposed that the polygons in Figure \ref{fig:polygon family} could tile the plane. Mathematically proving this is a non-trivial task, and their assertion on tiling is correct as we have established a tiling for each polygon. These tilings have been provided in the Appendix.

Claude 3 additionally suggested that all the polygons from Figure \ref{fig:polygon family} have an even number of right angles. This can be easily visually verified as correct. It is worth noting here that finding this property is also non-trivial, given that Claude 3 was only given the prompt and was not provided any visual data.


Whilst Claude 3 and Bing Copilot were able to provide new and interesting ideas, all of the LLMs also provided many ideas that were either incorrect or uninteresting (i.e. ``Every angle must be 90 or 270 degrees"). More specifically, Claude 3 suggested 7 properties of which 2 were interesting, 3 were uninteresting, and 2 were wrong; Bing Copilot suggested 11 properties of which 1 was interesting, 8 were uninteresting, and 2 were wrong; GPT-3.5-Turbo suggested 7 properties of which 5 were interesting and 2 were wrong. 

In some sense the wider performance from LLMs is rather like a mathematician in training. Not all suggestions made by human mathematicians initially would be interesting or correct, but may prompt further thinking. A full breakdown of the responses from LLMs, with conversational screenshots can be found in Appendix \ref{sec:Claude_response_shapes} to \ref{sec:Bing_response_shapes}.

\section{Limitations}

Due to the nature of our study, there are fundamental limitations to our work. Firstly, it is difficult to come up with complex and novel questions on which to test LLMs. As such, we have been unable to conduct tests on a large number of questions as compared to related literature  \cite{bubeck2023sparks}. In a similar vein, evaluating whether a response to a complex and novel question is correct is difficult and time consuming. Therefore we were not able to test using a large number of models or over a large number of iterations. Finally, we note that as LLMs are trained on large amounts of data it is difficult to guarantee that the questions we have asked are not in the training dataset, even if such a prospect is unlikely. This is a problem in other related literature, as well as the wider field of LLM evaluation. Nevertheless it is important to expose new observations on the capabilities of LLMs so that the wider body of knowledge accumulates. It is in this context that our findings are presented. 


\section{Conclusion}

Whilst the creativity and originality of AI is still a contentious issue both inside and outside of academia, there is growing evidence that AI can aid in many areas of research and design. There are also claims that AI can achieve unpublished results to some degree \cite{GuillaumeVerdon, BenBlaiszik}. We tested these claims on two problems that are, to the best of our knowledge, untouched by the existing literature. 

In both cases, Claude 3 was able to provide meaningful suggestions and contributions to the questions even if its answers were not perfect. For the decidable game, Claude 3 was able to find the optimal strategy for an odd number of spaces and elegantly use this result to correctly deduce winner of the case when there were 7 spaces. For the polygons task, meanwhile, Claude 3 made two novel suggestions for properties of these polygons. 

It is also worth noting that Bing Copilot was also able to provide some useful contributions as well, though certainly not to the same level as Claude 3, with Bing Copilot's most notable contribution being the suggestion that the polygons from the polygon question tiled the plane (which was also suggested by Claude 3). ChatGPT-3.5-Turbo, meanwhile, made few interesting correct suggestions. Notably, this observation combined with the above two observations indicates that newer models are more capable of providing novel insights into unseen problems.


A final interesting conclusion comes from noting that in the decidable game task, Claude 3 and ChatGPT-3.5-Turbo concluded that the game was Nim \cite{bouton1901nim}. Whilst this was incorrect, the suggestion prompted us to then check if an equivalence held via the Sprague-Grundy Theorem \cite{grundy1939mathematics}. This represents an interesting case wherein the LLM, whilst incorrect, was close enough to a correct statement to steer us towards the correct idea. The fact that these incorrect statements are able to provide insights on the correct statements is certainly exciting prospect, as it means that LLMs, in collaboration with humans, may be better than the sum of their parts in exposing new knowledge.

Within this paper we see examples of LLMs providing new insights into questions that are outside of the LLM's training data. Whilst these questions are perhaps simpler than many advanced research questions, that the LLM is able to provide new ideas is highly notable even if the outputs aren't perfect. 

We also see that current LLMs  have considerable limitations. Complex questions are difficult for LLMs to address, and much of the information generated is incorrect, reaffirming that advancing the next word prediction approach is something that the future architecture of LLMs needs to consider. This observation is in line with previous research \cite{bubeck2023sparks}. 
Nevertheless, we acknowledge that the LLMs are a new and evolving technology that appear to have considerable potential to support the generation of new results for research and innovation. 

\newpage
\bibliographystyle{plain}
\bibliography{bibliography.bib}

\newpage
\section{Appendix}

\subsection{An analysis of the responses finding the winner of a decidable game}

\subsubsection{Claude 3's response to the decidable game problem.}
\label{sec:Claude_response_game}

We begin our analysis of the game by exploring Claude 3's answer (see below):

\begin{figure}[h]
\centering
\includegraphics[scale = 0.5]{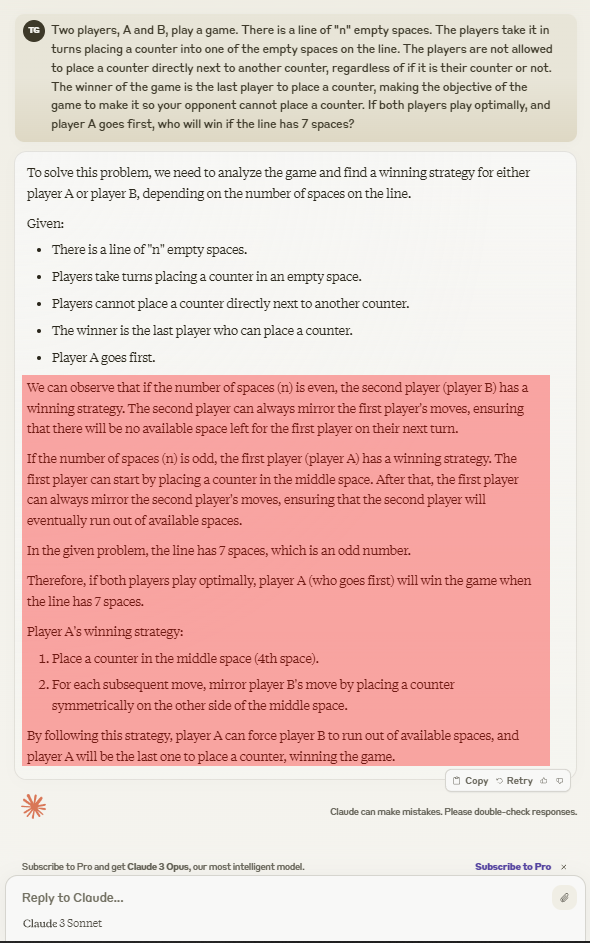}
\caption{Claude 3's attempt to find the winner.}
\label{fig:claude coin}
\end{figure}

The part where Claude 3 attempts to solve the problem in the above figure has been highlighted in red. We first note that the "winning strategy" for if the number of spaces is even is wrong, we find this via picking the number of empty spaces to be 2 and letting player A go first:

\begin{figure}[h]
\centering
\includegraphics[scale = 0.2]{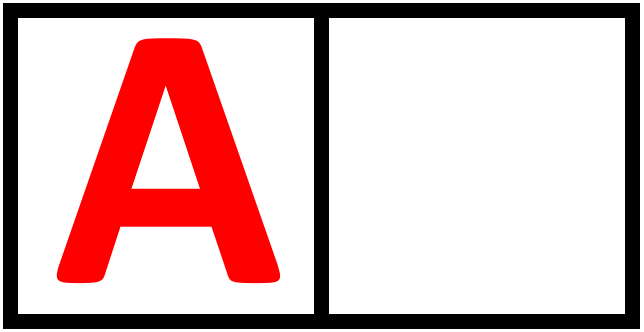}
\caption{The game when n = 2}
\label{fig:n is 2}
\end{figure}

In Figure \ref{fig:n is 2}, it can be seen that player B cannot place a counter and thus loses the game, but this means that player B does not have a winning strategy for when there are two empty spaces. But 2 is an even number, and Claude 3 claimed that player B has a winning strategy if there is an even number of spaces. This contradicts Claude 3's statement.

\subsubsection{Bing Copilot's response to the decidable game problem.}
\label{sec:Bing_response_game}

The strategy provided by Claude 3 in Figure \ref{fig:claude coin} for when there is an odd number of empty spaces is correct. This follows from the fact that placing a coin in the middle essentially splits the line of empty spaces into 2 lines of empty spaces of equal length, creating a symmetry that player A can utilise for the rest of the game. Claude 3 then correctly notes that this strategy can be used to allow player A to win when the number of empty spaces is 7. To our knowledge, this strategy and game do not currently exist in the academic literature.

The next response we consider is Bing Copilot, which can be seen below:

\begin{figure}[h]
\centering
\includegraphics[scale = 0.5]{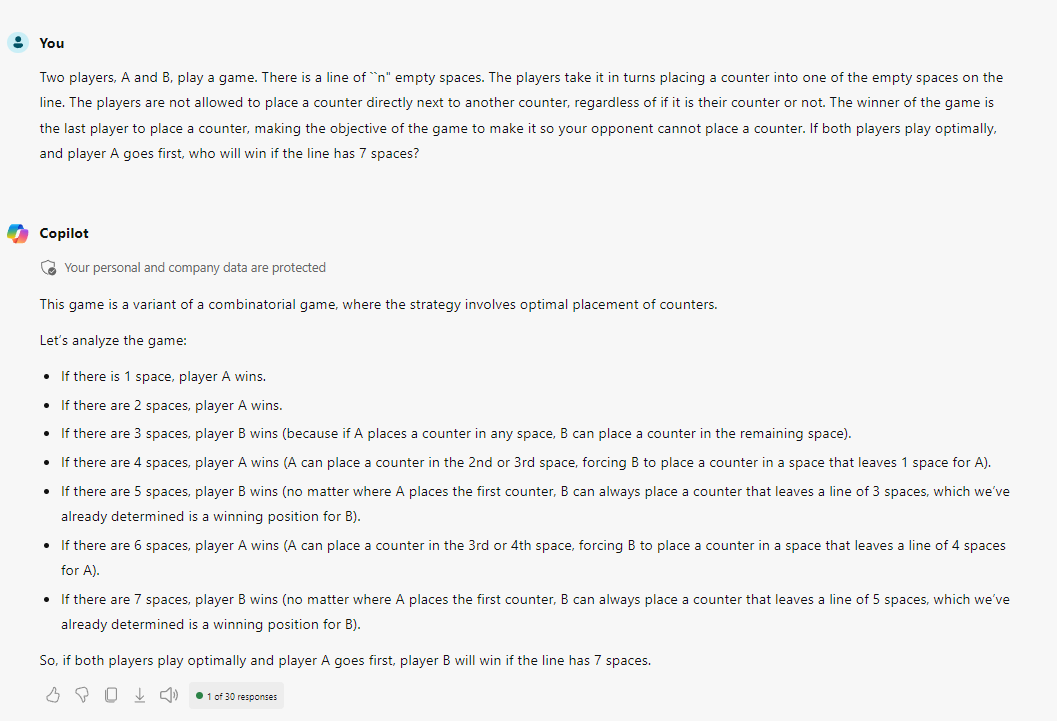}
\caption{Bing Copilots's attempt to find the winner}
\label{fig:bing coin}
\end{figure}

In Figure \ref{fig:bing coin}, Bing Copilot opts to approach the question by analysing specific cases. The cases for 1 empty spaces and 2 empty spaces are correct, however player A can win when there are 3 empty spaces by playing their first move in the centre of the row like so:

\begin{figure}[h]
\centering
\includegraphics[scale = 0.2]{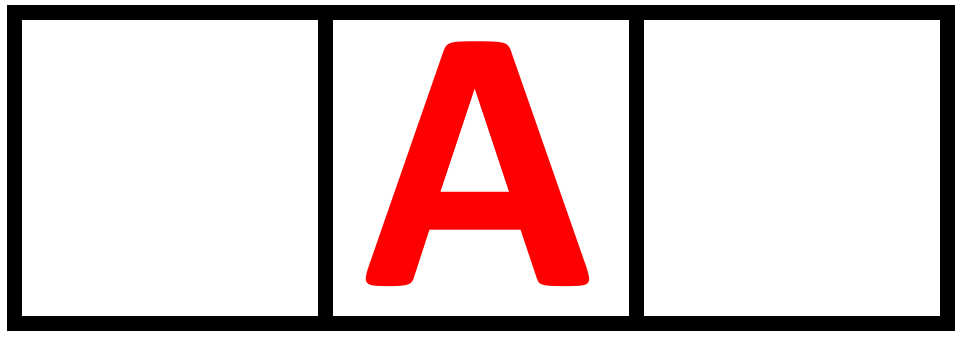}
\caption{The game when n = 3}
\label{fig:n is 3}
\end{figure}

Similarly, we can see a winning strategy for player A for when there are 5 empty spaces and 7 empty spaces via using the strategy that Claude 3 suggested in Figure \ref{fig:claude coin}. Bing Copilot's suggestions for 4 empty spaces and 6 empty spaces are also incorrect. The game when there are 4 empty spaces cannot be won by player A.

The case of 6 empty spaces, meanwhile, is winnable by player A, who can place their counter at the very end of the line (reducing the problem to the case with 4 empty spaces, which is unwinnable for player B).

\subsubsection{ChatGPT-3.5-Turbo's response to the decidable game problem.}
\label{sec:GPT_response_game}

Finally, we look at ChatGPT-3.5-Turbo's response (see below):

\begin{figure}[h]
\centering
\includegraphics[scale = 0.4]{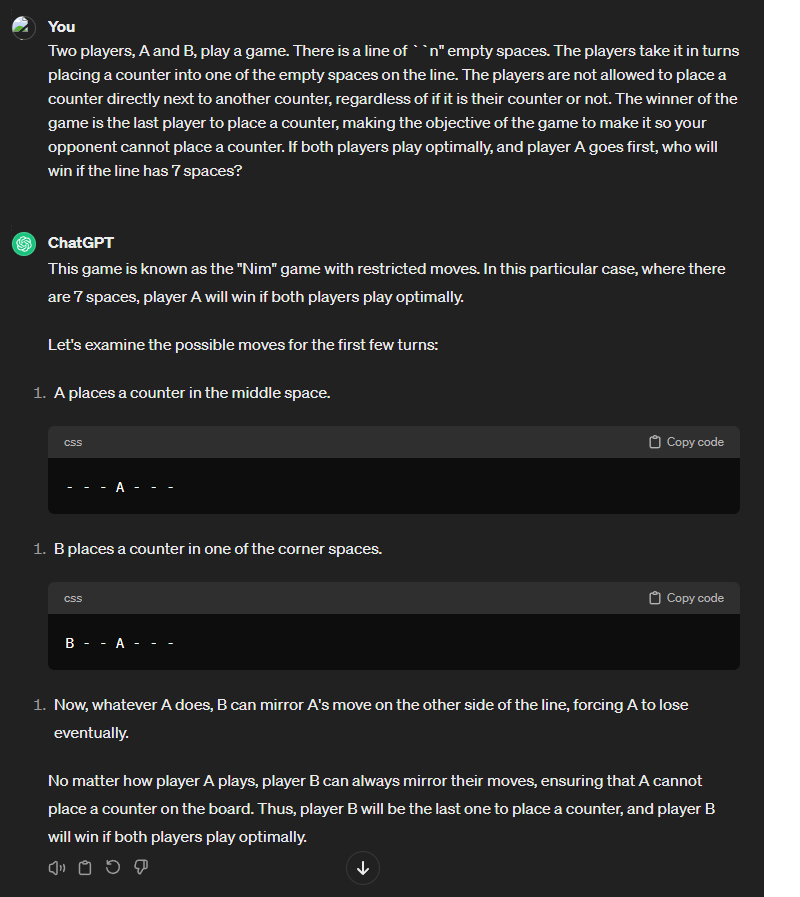}
\caption{ChatGPT-3.5-Turbo's attempt to find the winner.}
\label{fig:GPT coin}
\end{figure}

ChatGPT-3.5-Turbo attempts the question directly, and seems to get confused about turn orders. The strategy suggested by ChatGPT-3.5-Turbo is strikingly similar to the one suggested by Claude 3 in Figure \ref{fig:claude coin}. ChatGPT-3.5-Turbo also notably says that the game is a version of nim (Claude 3 also did this after being asked if it recognised the game, however this interaction has not been included to save space in the paper). The statement that the game is actually nim is not fully correct, but it is notable that the game can be proven to be equivalent to a version of nim via the Sprague-Grundy theorem \cite{grundy1939mathematics}.

\newpage
\subsection{An analysis of the responses to the polygons question}

\subsubsection{Claude 3's response to the polygon problem.}
\label{sec:Claude_response_shapes}

We begin our analysis of the polygons question by exploring Claude 3’s answer. Initially, Claude 3 was asked what properties it was aware of the polygons having:

\begin{figure}[h]
\centering
\includegraphics[scale = 0.6]{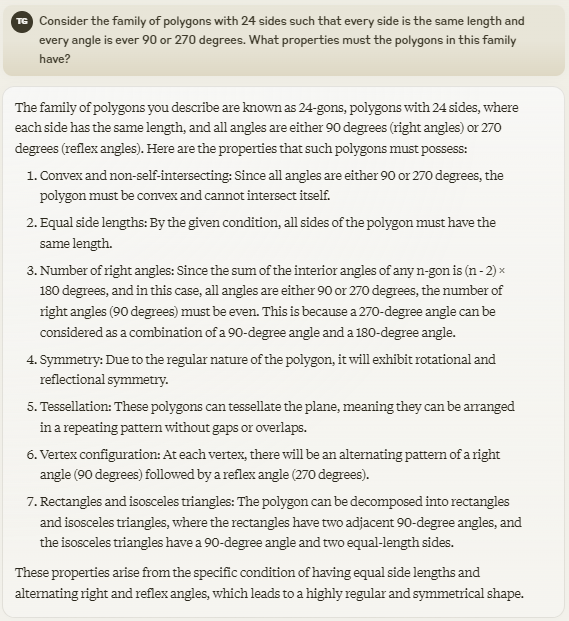}
\caption{Claude 3's first response to the polygon problem.}
\label{fig:Claude 3's polygon 1}
\end{figure}

The properties suggested by Claude 3 in Figure \ref{fig:Claude 3's polygon 1} are mosty non-trivial suggestions, however many of them are also unfortunately incorrect. The convexity suggestion is not correct, as a shape with an angle of 270 degrees is by definition not convex. Similarly, we can show symmetry to be incorrect as well via a counter example:

\begin{figure}[h]
\centering
\includegraphics[scale = 0.8]{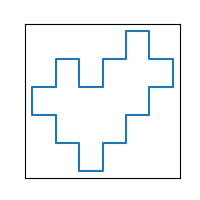}
\caption{A polygon meeting our rules that does not have symmetries}
\label{fig:counter example}
\end{figure}

The counter example in Figure \ref{fig:counter example} (see above) also shows that we will not always have a 90 degree angle followed by a 270 degree angle, which shows another one of Claude 3's suggestions to be incorrect.

If we consider only polygons that do not self-intersect and have no holes, we can see the third suggestion made by Claude 3 in Figure \ref{fig:Claude 3's polygon 1}, which states that there must be an even number of 90 degree angles within the polygons, is correct. We can see this by checking each of the 7 unique polygons for this number of sides directly (see below figure):

\begin{figure}[h]
\centering
\includegraphics[scale = 0.6]{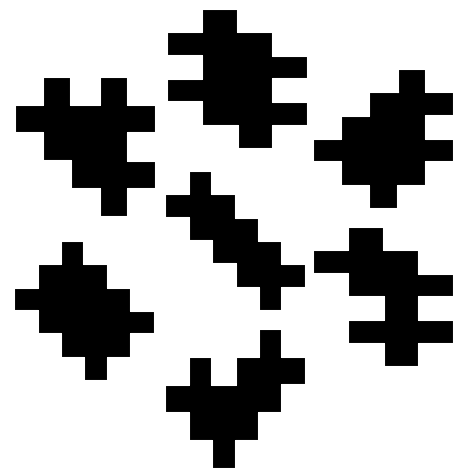}
\caption{All 7 polygons}
\label{fig:polygonsssss}
\end{figure}

The observation that all of the above polygons have an even number of right angles is surprising, both because it is a new result and because this statement, in the related field of polyominoes (polygons made up of squares), is not generally true. Consider, for example, the following polyomino with 5 right angles (see below figure):

\begin{figure}[h]
\centering
\includegraphics[scale = 0.4]{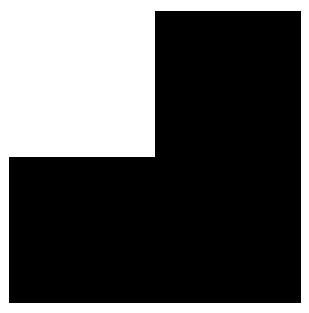}
\caption{A polyomino with 5 right angles}
\label{fig:tetris}
\end{figure}

When asked to justify the claim that all of the polygons in Figure \ref{fig:polygonsssss} had an even number of right angles (note here that Claude 3 was never provided an image of any of the polygons in Figure \ref{fig:polygonsssss} during this conversation), Claude 3 was not able to give any correct reasoning.

We next consider the statement by Claude 3 that the polygons in Figure \ref{fig:polygonsssss} can be split into rectangles and triangles. This is correct, and can be seen by noting that the polygons in Figure \ref{fig:polygonsssss} are a subset of the polyominos, and thus can be decomposed into squares (which are rectangles). The squares can then, in turn, be decomposed into isosceles triangles via drawing a line between two diagonally opposite points in the square.

Finally, we look into Claude 3's suggestion that the polygons in Figure \ref{fig:polygonsssss} tile the plane. This is difficult to rigorously prove, however we can see a potential tiling for each polygon on the next page (readers are encouraged to note there that at no point in this conversation was Claude 3 shown the polygons from Figure \ref{fig:polygonsssss}). See the figure on the next page for a potential tiling for each polygon.

We then asked Claude 3 if it could prove that the polygons from Figure \ref{fig:polygonsssss} can tesselate the plane. The proof Claude 3 provided was incorrect. When Claude 3 was asked how many polygons exist in this family, it suggested that there were 2 with very poor reasoning for why this should be true.

\newpage
\begin{figure}[h]
\centering
\includegraphics[scale = 0.134]{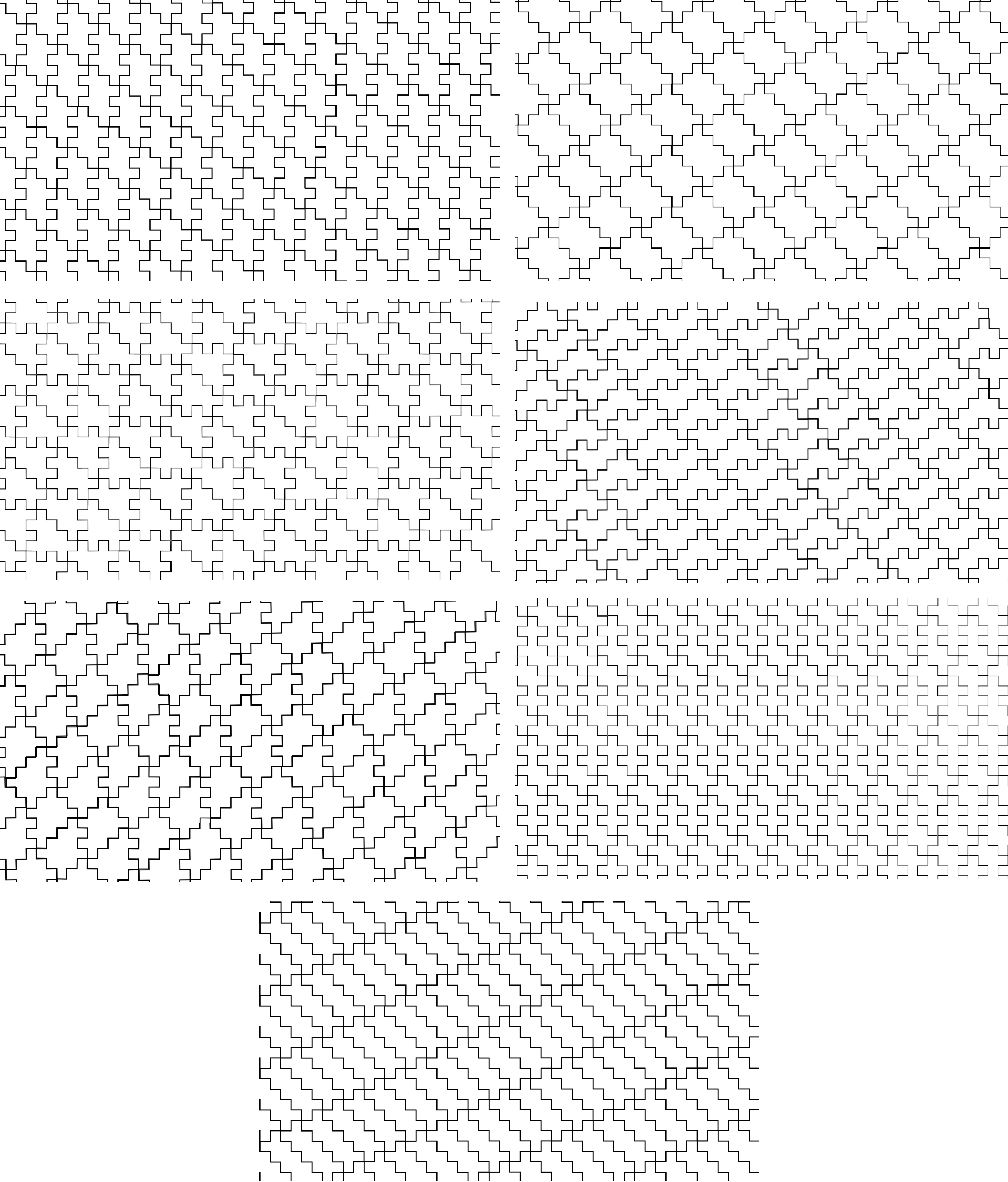}
\caption{The potential tiling for each polygon.}
\label{fig:tilings}
\end{figure}
\newpage

\subsubsection{ChatGPT-3.5-Turbo's response to the polygon problem.}
\label{sec:GPT_response_shapes}

Next, we asked Chat-GPT-3.5-Turbo about the properties of polygons in this family:

\begin{figure}[!ht]
\centering
\includegraphics[scale = 0.3]{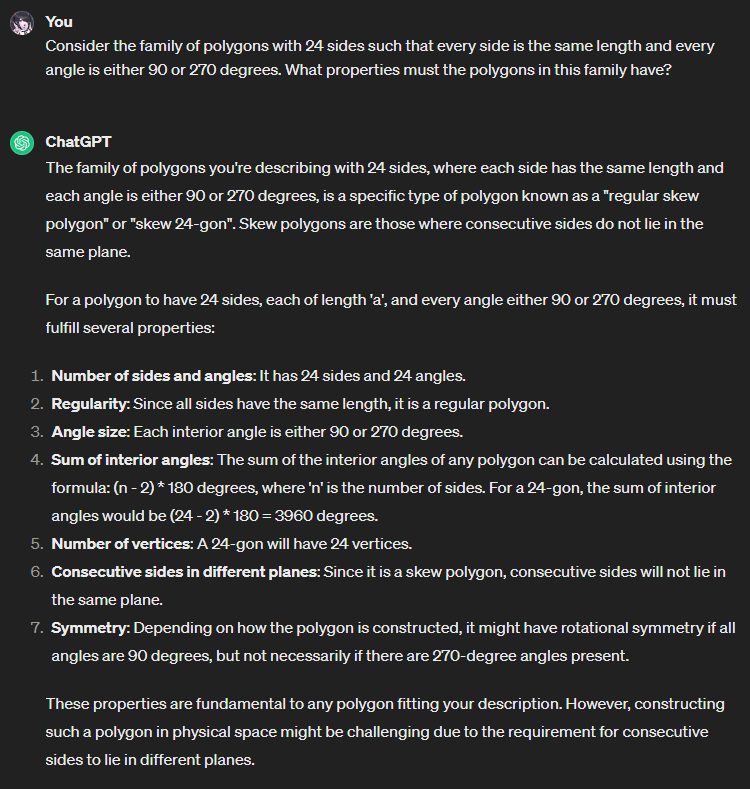}
\caption{Chat-GPT-3.5-Turbo's first response to the polygon problem.}
\label{fig:GPT polygon 1}
\end{figure}

ChatGPT-3.5-Turbo suggests that the polygon family here is a family of skew polygons, meaning that the polygon cannot be 2D. This was unexpected, but not disallowed by the rules set in place. Suggestions 1, 3, 4, and 5 from figure \ref{fig:GPT polygon 1} are all obvious from the initial question. The claim of regularity is not true (as regular implies equilateral AND equiangular). If you have two consecutive sides in a skew polygon they must be in the same plane making suggestion 6 incorrect. Finally, we note that whilst some polygons from this family are symmetrical, not all of them are - and if we consider skew polygons, even if there is only 90 degree angles in the polygon we may not necessarily have rotational symmetry.

After this, we asked ChatGPT-3.5-Turbo if it knew any less obvious properties, however ChatGPT-3.5-Turbo wasn't able to give any interesting ideas, instead saying that the polygon family may or may not have certain properties. Finally, we asked how many polygons existed in this family:

\begin{figure}[h]
\centering
\includegraphics[scale = 0.3]{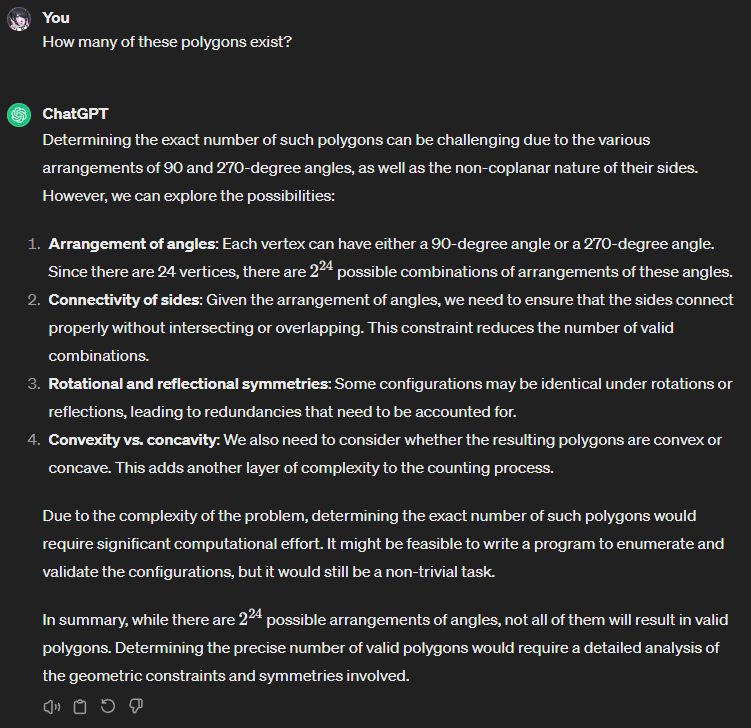}
\caption{Chat-GPT-3.5-Turbo's third response to the polygon problem.}
\label{fig:GPT polygon 2}
\end{figure}

Whilst the reasoning given in Figure \ref{fig:GPT polygon 2} would usually be correct, when considering skew polygons there are more arrangements not accounted for by ChatGPT-3.5-Turbo.

\subsubsection{Bing Copilot's response to the polygon problem.}
\label{sec:Bing_response_shapes}

The final model we tested on this task was Bing Copilot. Initially, we asked Bing Copilot what properties this family of polygons have:

\begin{figure}[h]
\centering
\includegraphics[scale = 0.34]{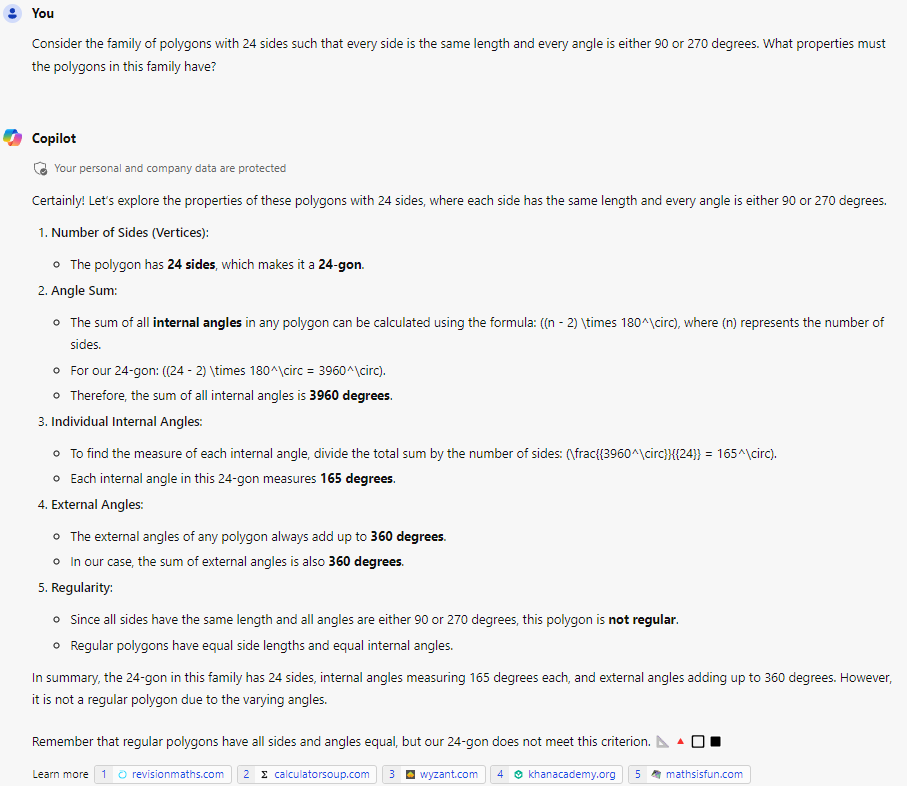}
\caption{Bing Copilot's first response to the polygon problem.}
\label{fig:Bing Copilots's polygon 1}
\end{figure}

Each one of the suggested properties was, unfortunately, properties of polygons in general or just restatements of the problem it was given, therefore, we asked Bing Copilot if it knew of any less obvious properties:

\begin{figure}[h]
\centering
\includegraphics[scale = 0.34]{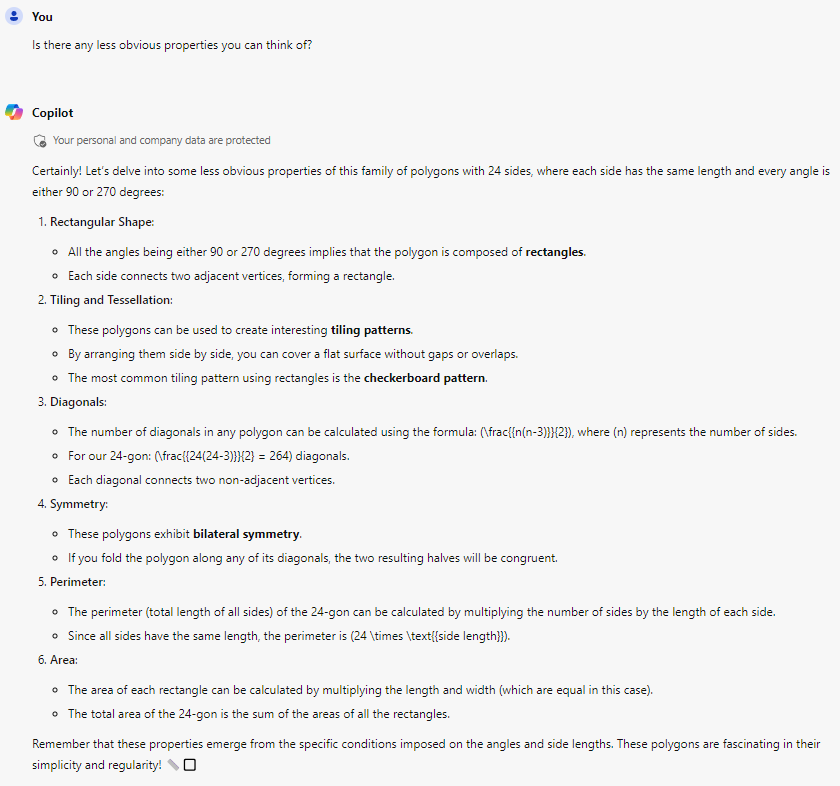}
\caption{Bing Copilot's second response to the polygon problem.}
\label{fig:Bing Copilots's polygon 2}
\end{figure}

Interestingly Bing Copilot, similarly to Claude 3, suggests that this family of polygons may tile the plane (a statement we have already seen to be correct). The rest of Bing Copilot's suggestions - meanwhile - were unfortunately incorrect. Asking Bing Copilot if it could prove that the polygons could tile the plane resulted in an explanation of the history of plane tiling which, whilst interesting, was not what was requested.

Finally, we asked Bing Copilot how many of these polygons existed in this family. Here, Bing Copilot re-iterated some properties suggested earlier, and then concluded that the polygons in this family were not regular polygons.

\end{document}